\documentclass{article}

\usepackage{arxiv}

\usepackage[utf8]{inputenc} 
\usepackage[T1]{fontenc}    
\usepackage{hyperref}       
\usepackage{url}            
\usepackage{booktabs}       
\usepackage{amsfonts}       
\usepackage{nicefrac}
\usepackage{microtype}
\usepackage{cleveref}
\usepackage{lipsum}
\usepackage{graphicx}
\usepackage{doi}
\usepackage{caption}
\usepackage{subcaption}
\usepackage{mwe}
\usepackage{hyperref}

\title{Bugs in the Data: \\ How ImageNet Misrepresents Biodiversity}
\author {
    Alexandra Sasha Luccioni\thanks{Equal contribution}\\
    Hugging Face \\
	\texttt{sasha.luccioni@huggingface.co}\vspace{-0.3in}\\
	\And
    David Rolnick\footnotemark[1] \\
    McGill University, Mila \\
    \texttt{drolnick@cs.mcgill.ca\vspace{-0.3in}}
}
\date{}

\renewcommand{\shorttitle}{Bugs in the Data: How ImageNet Misrepresents Biodiversity}

\begin{document}

\maketitle

\begin{abstract}
\emph{ImageNet-1k} is a dataset often used for benchmarking machine learning (ML) models and evaluating tasks such as image recognition and object detection. Wild animals make up 27\% of ImageNet-1k but, unlike classes representing people and objects, these data have not been closely scrutinized. In the current paper, we analyze the 13,450 images from 269 classes that represent wild animals in the ImageNet-1k validation set, with the participation of expert ecologists. We find that many of the classes are ill-defined or overlapping, and that 12\% of the images are incorrectly labeled, with some classes having $>$90\% of images incorrect. We also find that both the wildlife-related labels and images included in ImageNet-1k present significant geographical and cultural biases, as well as ambiguities such as artificial animals, multiple species in the same image, or the presence of humans. Our findings highlight serious issues with the extensive use of this dataset for evaluating ML systems, the use of such algorithms in wildlife-related tasks, and more broadly the ways in which ML datasets are commonly created and curated.
\end{abstract}

\section{Introduction} \label{sec:introduction}

Datasets are a crucial part of the development, evaluation and eventual deployment of machine learning (ML) systems. Large datasets such as ImageNet~\cite{deng2009imagenet} and MS-COCO~\cite{lin2014microsoft} have been used for years both as inputs for training ML models for downstream tasks and as benchmarks for evaluating model efficacy. There are many choices involved implicitly or explicitly in the creation of a labeled dataset, including how classes are defined as well as the way datapoints are assigned to them. Recent research has placed increased scrutiny on many popular datasets, uncovering biases~\cite{prabhu2020large}, duplicates~\cite{barz2020we} and problematic class definitions~\cite{crawford2019excavating}. Such dataset shortcomings can have serious impacts on the behavior of the models they are used to train, and can also make it difficult to evaluate the true effectiveness of ML systems, since metrics such as accuracy are impacted by biases or inaccuracies in the evaluation set.

The ImageNet Large Scale Visual Recognition Challenge (ILSVRC) dataset, commonly called \emph{ImageNet-1k}, \cite{russakovsky2015imagenet} is a subset of the full ImageNet dataset~\cite{deng2009imagenet} that is used widely in categorical object recognition (with 32k citations to date). More than a quarter of the dataset consists of images of wild animals. However, no wildlife experts were consulted in creating the dataset, and the choice of classes and images has never been analyzed, despite the considerable role that these data have ultimately played in shaping widely used ML models.

Here, we examine the 269 classes of wild animals in ImageNet-1k, working with expert ecologists to evaluate both the class definitions and the images included in each class. Our main findings are the following:
\begin{itemize}
    \item Over 12\% of the wildlife images are incorrectly labeled, with some classes having over 90\% incorrect labels.
    \item Many images are problematic in other ways, including the presence of humans (6.7\% of images), multiple types of animals in the same image (4.5\%), images too blurry for identification (2.2\%), and images of fake animals such as stuffed toys (1.4\%).
    \item Both the class definitions and images are strongly biased towards the U.S.~and Europe, vastly under-representing biodiversity from other geographies.
    \item Many classes in the dataset are not clearly defined (11.5\% of classes) or overlap with other classes (11.9\%).
\end{itemize}

By leveraging the concrete nature of wildlife images, where individual species (including their geographic provenance) can be readily identified, we provide a novel window into the many weaknesses of ImageNet-1k as a benchmark dataset. Our findings also raise significant concerns about various real-world ML applications where ImageNet-1k is used, as well as highlighting the fundamental importance of domain expertise in crafting ML benchmark datasets.

\begin{figure}[ht!]
\centering
\begin{subfigure}{.23\columnwidth}
\includegraphics[width=\columnwidth]{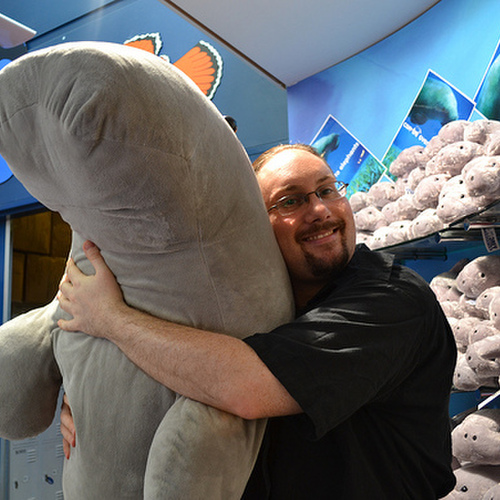}%
\end{subfigure}\quad%
\begin{subfigure}{.23\columnwidth}
\includegraphics[width=\columnwidth]{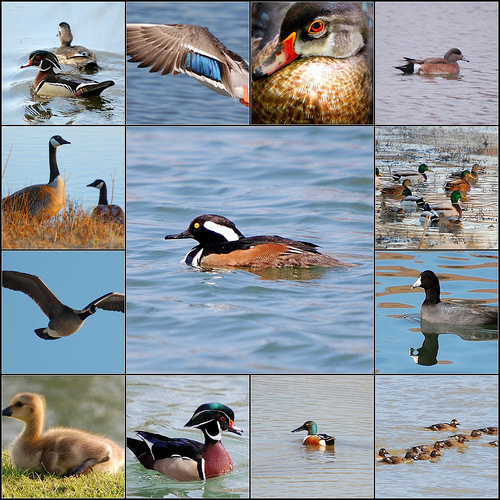}
\end{subfigure}\quad
\begin{subfigure}{.23\columnwidth}
\includegraphics[width=\columnwidth]{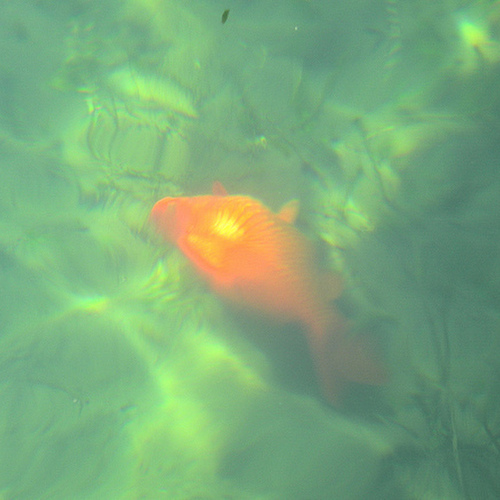}
\end{subfigure}\quad
\par\medskip 
\begin{subfigure}{.23\columnwidth}
\includegraphics[width=\columnwidth]{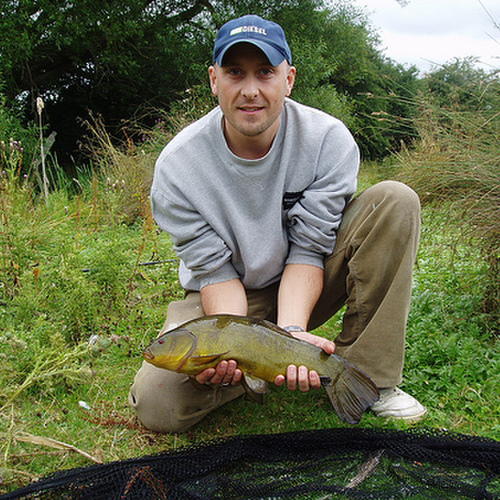}%
\end{subfigure}\quad%
\begin{subfigure}{.23\columnwidth}
\includegraphics[width=\columnwidth]{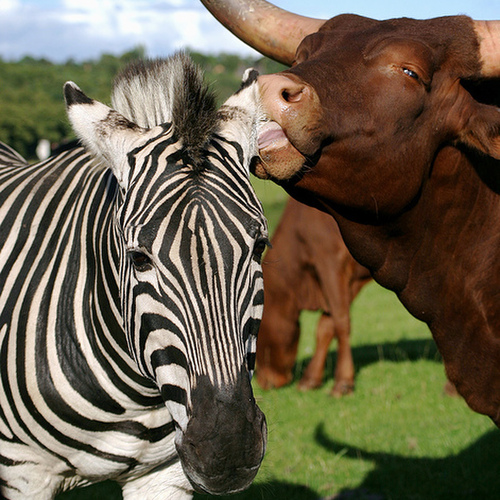}
\end{subfigure}\quad
\begin{subfigure}{.23\columnwidth}
\includegraphics[width=\columnwidth]{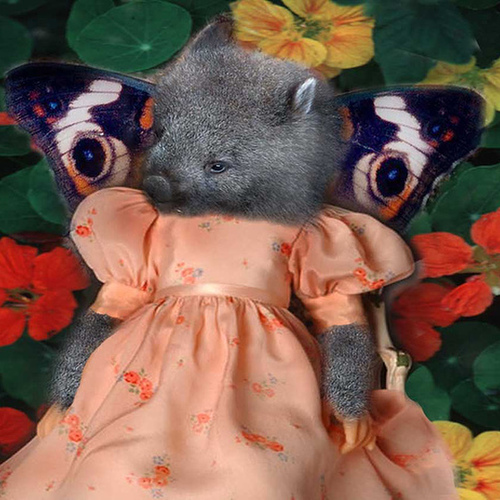}
\end{subfigure}\quad
\caption{Examples of images of animals from ImageNet-1k, including artificial animals (top left), collages (top right) (here, the correct species ``American coot'' occurs only in one of the side panels of the image), blurry images (center left), images with humans (center right), images with multiple species of animals (bottom left) and images that are bizarre in other ways (bottom right: a wombat in a dress with butterfly wings and human hands).\vspace{-0.1in}}
\label{fig:examples}
\end{figure}

\section{Related Work} \label{sec:related_work}

Our work is relevant to several research directions, ranging from empirical studies on the impact of label errors on model performance to analyses of the values embedded in ML datasets. We briefly summarize work in these fields in the paragraphs below. 

\paragraph{A brief history of ImageNet.} Since the creation in 2009 of the original ImageNet dataset \cite{deng2009imagenet}, containing 21,000 classes, it has gone through numerous iterations, including the 1k version created for the ImageNet Large Scale Visual Recognition Challenge in 2010~\cite{russakovsky2015imagenet}, commonly called \emph{ImageNet-1k}. In recent years, following an in-depth study by the Excavating AI project~\cite{crawford2019excavating}, an updated official version of the 21K dataset was created, aiming to filter out problematic person categories and to balance their distribution~\cite{yang2020towards}. In 2021, a subset of the authors of the original ImageNet paper proposed a new version of the full dataset, using facial recognition techniques to blur the faces of people in the images, showing that it does not impact model performance~\cite{yang2021study}. 

\paragraph{Analyses of ImageNet.} There have been numerous studies of ImageNet from different angles, ranging from the quality of its labels~\cite{beyer2020we,northcutt2021pervasive,tsipras2020imagenet} to the biases that it reflects~\cite{crawford2019excavating,prabhu2020large}. There have also been attempts to replicate its data collection process~\cite{recht2019imagenet} and to propose a new validation set for ImageNet, meant to better present ``real-life'' images~\cite{barbu2019objectnet}. To date, these studies have focused on the the ``people'' and ``object'' classes of ImageNet, with the notable exception of a 2015 study that looked at the bird classes in ImageNet, hypothesizing class error rates of at least 4\%~\cite{van2015building}. We build on this work, examining all of the wildlife classes of ImageNet-1k and not only calculating the class error rates (which we find are much higher for other animals than for birds), but also analyzing the biases and values implicit in the class definitions and choice of images. 

\paragraph{Work on wildlife image datasets.} ImageNet is far from the only large dataset containing images of wildlife; in fact, many researchers have created diverse datasets specifically focused on wildlife images in the last decade, often working hand-in-glove with biodiversity experts and citizen scientists. For instance, the eBird dataset was created at the same time as ImageNet and contains nearly 3 million images of birds, collected and labelled by citizen scientists~\cite{sullivan2009ebird}, which have been extensively utilized and studied~\cite{van2017devil}. \cite{van2015building} proposed \emph{NABirds}, a dataset containing almost 50,000 images of North American birds spanning 555 classes. The iWildCam Competition is another notable example of extensive biodiversity datasets, leveraging camera traps to gather hundreds of thousands of images across the world~\cite{beery2021iwildcam,beery2019iwildcam}; these datasets have been used to host competitions, establish benchmarks, and assess model generalization to novel ecosystems and species~\cite{beery2018recognition}, but remain underutilized by the ML community at large. Finally, one of the most diverse and extensive biodiversity datasets is iNaturalist, which spans 859,000 images from over 5,000 species of plants and animals~\cite{van2018inaturalist} and has been used to develop highly popular ML-powered species identification mobile applications such as Seek~\cite{seek}.

\paragraph{Values and biases embedded in datasets.} Recent years have also seen several deep dives into the provenance and values latent in ML datasets. For instance, Crawford and Panglen's ``archaeology" of several ML datasets including ImageNet has exposed the racist and misogynist assumptions and values that are buried in them~\cite{crawford2019excavating}. This has been built upon by further scholarship around the ethical and political dimensions of dataset curation and construction~\cite{ denton2020bringing,prabhu2020large} and the values that are embedded in datasets~\cite{scheuerman2021datasets}. Complementary work has been carried out to study how datasets are created and used by the community, finding that there is a concentration of power in terms of institutions that create, which results in influence of ML research and practice~\cite{koch2021reduced}. This work has inspired our own, since we find that these values and practices have also impacted the representation of nature and biodiversity in ML datasets, and more specifically ImageNet-1k, which is the focus of our study.

\section{Methodology} \label{sec:methodology}

In this section, we detail the approach that we used for systematically evaluating the images of wildlife in ImageNet. 

\subsection{Selecting Images of Wildlife} \label{sec:sampling}

There are 398 classes representing animals in ImageNet-1k, which is over a third of the 1000 classes in the overall dataset. Of these, 269 represent wild extant animals, whereas 127 are breeds of domestic animals such as dogs, cats and chickens, and 2 represent prehistoric extinct animals (the trilobite and triceratops). In the present study, we focus on wild animals, given that domestic breeds' presence depends wholly on cohabitation with humans, and many of the same domestic breeds are now present across the globe, notwithstanding whether they were originally from a specific region. We also exclude long extinct species, since it is clearly impossible to have photos of these creatures in the wild (which entails that all of the images of these species are artists' representations or photographs of fossils). We separate these 269 classes of wild animals into 7 taxonomic groups, as shown in Table \ref{fig:nature-categories}. 

\begin{table}[ht]
\begin{center}
\begin{tabular}{l|c|c}
\textbf{Taxonomic group}   & \textbf{\#~classes} & \textbf{\#~images} \\ \hline
Amphibian    & 8  & 400 \\
Bird        & 57    & 2850 \\
Fish       & 16  & 800   \\
Mammal    &  93   & 4650    \\
Marine Invertebrate & 20  & 1000    \\
Reptile    & 35    & 1750   \\
Terrestrial Invertebrate & 40  & 2000    \\ \hline
Total        & 269      & 13450        
\end{tabular}\vspace{0.2in}
\caption{Summary of wildlife images in ImageNet-1k, broken down according to taxonomic group (here and throughout the paper, ``reptile'' refers to non-avian reptiles). There are 50 images per class in the validation set.\vspace{-0.1in}}
\label{fig:nature-categories}
\end{center}
\end{table}

\subsection{Domain Expert Validation} \label{sec:experts}

In order to validate both the class labels used in ImageNet-1k as well as the images from the validation set tagged with these labels, we recruited 20 expert annotators with expertise across different taxonomic groups. We use images from the validation set as this is designed to represent a representative subset of the full dataset and is of a tractable size ($\sim$13k images) to permit careful annotation by experts. Each of the expert annotators had at minimum graduate-level education in the domain in question -- for instance, the images of primates were annotated by two postdoctoral researchers in primate biology and one zoo keeper specializing in primates. Further details on our annotation process are given in Appendix \ref{app:annotation} and we provide the summarized annotations in our \href{https://github.com/RolnickLab/ImageNetBiodiversity}{GitHub repository}.

\subsubsection{Class Validation}

We asked our expert annotators the following questions for each class in their domain of expertise:

\begin{itemize}
    \item \textbf{Clarity:} Does the category clearly define a species or group of species? \texttt{[Yes/No]}
    \item \textbf{Species:} Is the category a single species?  \texttt{[Yes/No]}
    \item \textbf{Overlap:} Does the category overlap with another category in the list? If so, which category does it overlap with?
\end{itemize}

These questions aimed to validate both the scientific relevance and clarity of the classes in ImageNet-1k, and to estimate the degree of overlap between the classes, given that the 1000 synsets that were selected for ImageNet-1k are not meant to have any overlap between them. Furthermore, most of the classes have multiple labels spanning both formal and colloquial terms (e.g.~``platypus, duckbill, duckbilled platypus, duck-billed platypus, \textit{Ornithorhynchus anatinus}''), inherited from the original WordNet hierarchy~\cite{miller1995wordnet}, and it is important to validate the correspondence of these different terms from a scientific perspective. 

\subsubsection{Image Validation}

We also asked our expert annotators questions for each image in their domain of expertise:

\begin{itemize}
    \item \textbf{Correct ID}: To the extent of your knowledge, does the label correspond to the animal that is shown on the image? \texttt{[Yes/No/Maybe]}
     \item \textbf{Humans present}: Are there humans present in the image? \texttt{[Yes/No]}
     \item \textbf{Other animals present}: Are there other (non-human) animals present in the image? \texttt{[Yes/No]}
     \item \textbf{Not real animal}: Is the animal in the image an illustration, sculpture, or other representation of the animal? \texttt{[Yes/No]}
     \item \textbf{Blurry image}: Is the image too blurry or low quality to allow identification? \texttt{[Yes/No]}
     \item \textbf{Image collage}: Is the image composed of several images (either of the same animal or different ones)? \texttt{[Yes/No]}
\end{itemize}

See Figure~\ref{fig:examples} for examples of some of the failure modes indicated in the questions above.
 
\subsection{Species-level Re-annotation} \label{sec:bird-reannotate}

In order to better examine the biases and inaccuracies in the dataset, we also asked experts in bird classification to re-annotate all of the 2950 images from the 57 classes of birds, identifying each of the images to species level where possible. This experiment was designed to provide information about (i) the true species if the image was incorrectly classified, (ii) the individual species if the class label was vague (e.g.~``kite'' can refer to many different species of birds found in different parts of the world). We focused on the birds because birds can often be identified to the level of species based on images alone (as opposed to insects, for example).

\section{Findings} \label{sec:results}

We identify three categories of problems with wildlife-related data in ImageNet: (1) inconsistencies and biases in the choice and definition of classes, (2) inaccuracies in the images chosen to represent those classes, and (3) images that are not strictly inaccurate but exhibit other kinds of biases.

\subsection{Inconsistencies and Biases in Class Definitions}
\label{subsec:classbias}

Before even considering the images within the ImageNet-1k dataset, it is worth analyzing the set of classes. We find that many classes are unclearly defined or even overlap with other classes, resulting in ambiguity about the correct label for images. Beyond this, the classes chosen reflect biases related to geography as well as to cultural knowledge.
\begin{table}[ht]
\begin{center}
\begin{tabular}{l|c|c|c}
\textbf{Taxonomic group}                         & \multicolumn{1}{l}{\textbf{\begin{tabular}[c]{@{}l@{}}Class \\ Overlap\end{tabular}}} & \multicolumn{1}{l}{\textbf{\begin{tabular}[c]{@{}l@{}}Single \\ species\end{tabular}}} & \multicolumn{1}{l}{\textbf{\begin{tabular}[c]{@{}l@{}}Unclear \\ definition\end{tabular}}} \\ \hline
Amphibian   & 0.0\%     & 75.0\%    & 	25.0\%       \\
Bird    & 8.8\%    & 43.9\%  & 19.3\% \\
Fish   & 12.5\%    & 56.3\%  & 6.3\%  \\
Mammal  & 15.1\%  & 65.6\%  & 2.2\%   \\
Marine Invertebrate  & 20.0\%  & 20.0\%  & 20.0\%   \\
Reptile   & 14.3\%   & 54.3\%     & 20.0\%  \\
Terrestrial Invert. & 5.0\%   & 15.0\%  & 10.0\% \\ \hline
Overall & 11.9\% & 48.3\% & 11.5\% 
\end{tabular}\vspace{0.2in}
\caption{Percentages of classes with certain attributes: overlap with other classes (highly problematic), denoting a single species (e.g.~``coho salmon'') rather than a group of species (e.g.~``scorpion''), and having a clear definition.\vspace{-0.1in}}
\label{table:class}
\end{center}
\end{table}

\paragraph{Unclear classes.} As shown in Table \ref{table:class}, our expert annotators found 11.5\% of the wildlife classes in ImageNet-1k to be unclear, where they were not certain what types of wildlife were included. In some cases, this results from seemingly contradictory descriptors, such as ```hognose snake, puff adder, sand viper,'' since ``sand viper'' and ``hognose snake'' refer to different species. In other cases, lack of clarity results from terms (e.g.~``cricket'') with no single accepted definition. Some terms, such as the bird class ``kite,'' can be defined in different ways based on dialect, introducing geography-mediated ambiguity based on how the class name is interpreted.

\paragraph{Overlapping classes.} Even more concerning, our annotators found that 11.9\% of the wildlife classes used in ImageNet are actually overlapping (see Table \ref{table:class}), resulting in ambiguity about whether images belong in one or another. In some cases, this arises from vague class definitions; for example, the ImageNet class ``hognose snake, puff adder, sand viper'' is poorly defined and could overlap with the class ``horned viper, cerastes, sand viper, horned asp, Cerastes cornutus.'' In other cases, a generic class is used alongside a more specific one; for example, the class ``meerkat'' refers to a type of mongoose, even though ``mongoose'' is another ImageNet-1k class. Such behavior directly contradicts the intentions stated in~\cite{russakovsky2015imagenet}, which indicates that \textit{``the 1000 synsets are selected such that there is no overlap between synsets: for any synsets $i$ and $j$, $i$ is not an ancestor of $j$ in the ImageNet hierarchy.''} Also, several instances of confusion arise from one of the classes being essentially human-defined, rather than a taxonomic category, e.g.~the class ``tusker'' (referring to elephants with long tusks), which conflicts with the class ``African elephant,'' since most tuskers are African elephants.

\begin{figure}[ht]
    \centering
    \includegraphics[width=0.65\linewidth]{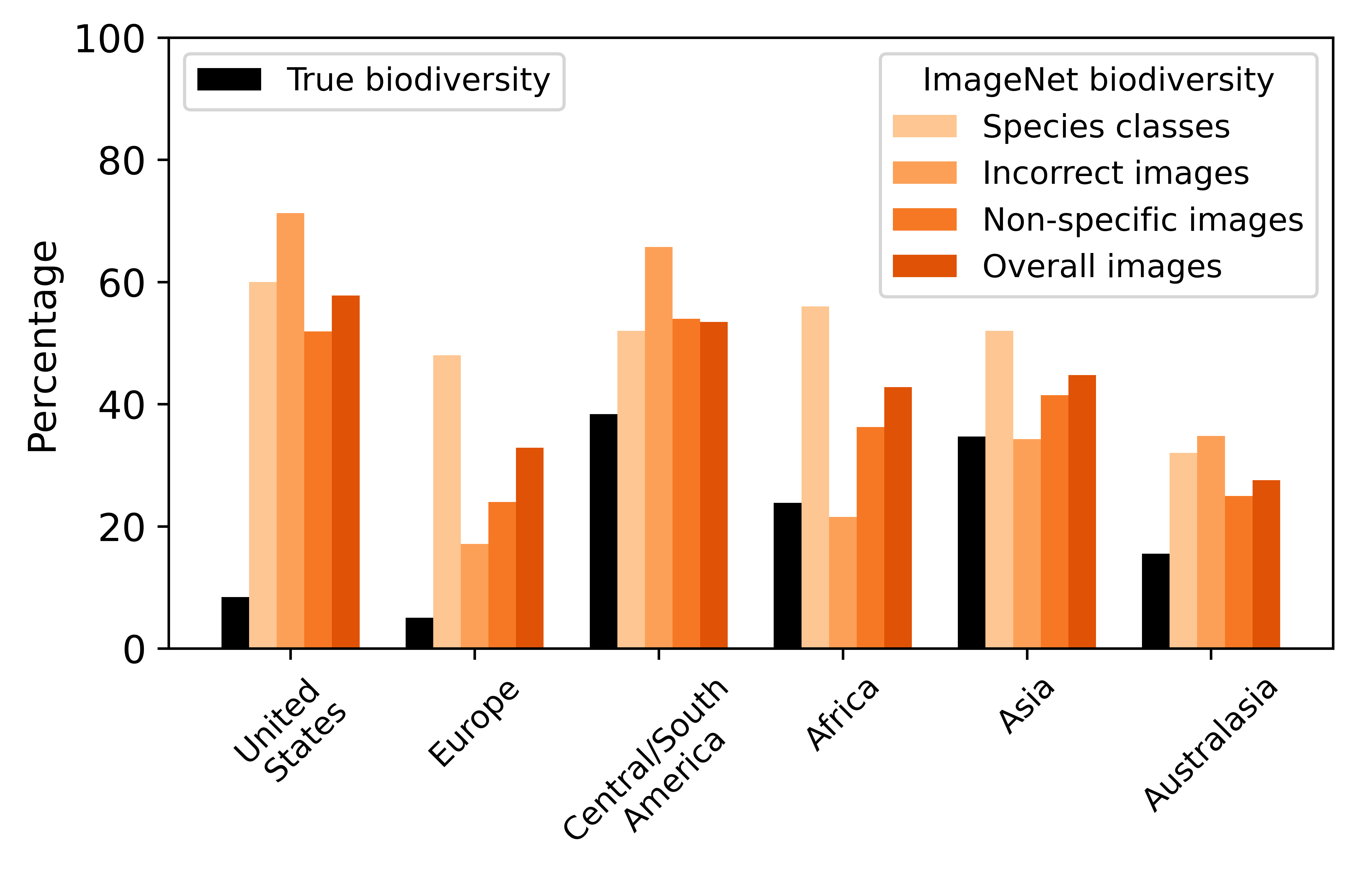}
    \caption{Geographic biases illustrated via comparison of the true biodiversity of birds and the biodiversity of birds within ImageNet-1k. ``True biodiversity'' shows the percentages of global bird species present in various geographic regions (note that these percentages sum to greater than 100\% since a single species may be present in multiple geographic regions). ``Species classes'' analyzes the geographic distribution of those classes of ImageNet-1k referring to individual species of birds, showing that the U.S.~and Europe are vastly overrepresented. ``Incorrect images'' analyzes where incorrectly identified birds are actually from, exhibiting a strong bias towards the U.S. ``Non-specific images'' analyzes which bird species are chosen to represent classes referring to multiple species (e.g.~62\% of ``jay'' images depict the blue jay, a U.S.~species). ``Overall images'' analyzes the overall distribution of bird species across all ImageNet-1k images.\vspace{-0.1in}}
    \label{fig:birdspecies}
\end{figure}
\paragraph{Geographic bias.} There are significant geographic biases in the classes represented in the dataset. For example, out of the sixty species of eagle that exist globally (many of which are culturally important birds in their regions) only the bald eagle (\textit{Haliaeetus leucocephalus}) is featured in the list of ImageNet-1k classes. Not coincidentally, it is the national bird of the United States -- indeed, out of the 25 species-specific classes of birds in ImageNet, 60\% are present in the U.S.~and 48\% are present in Europe, even though only 8.4\% and 5.0\% percent of bird species in the world are present in the U.S.~or Europe, respectively (see Figure \ref{fig:birdspecies}).\footnote{For the purposes of this analysis, we used bird lists from Avibase \cite{avibase}, neglecting vagrants and accidentals.} By contrast, bird diversity in Central and South America, Africa, Asia, and Australasia is severely underrepresented.

\paragraph{Cultural bias.} Linked to geographical bias in the choice of classes is cultural bias. By this, we refer to the selection of animals that are well-known or important culturally, rather than either (i) reflecting the balance of biodiversity, or (ii) reflecting the animals most widely seen in practice. Thus, for example, there are 93 classes referring to mammals in the dataset, but only 27 classes for insects, even though there are about six thousand known species of mammals in the world, compared to a million known species of insects~\cite{mora2011many} -- this reflects the relative importance of mammals culturally as compared to insects. Furthermore, within the ImageNet-1k classes for insects, only one refers to an individual species (the monarch butterfly, \textit{Danaus plexippus}, probably the most widely known butterfly in the U.S.), while all the other classes are more high-level and do not refer to a single species (``ant,'' ``dragonfly,'' etc.).  Implicit in the notion of cultural bias is the set of cultures within which cultural relevance is defined; as noted above, the set of relevant cultures is Americentric and Eurocentric.

\begin{figure*}[ht]
\begin{center}
\includegraphics[width=0.42\linewidth]{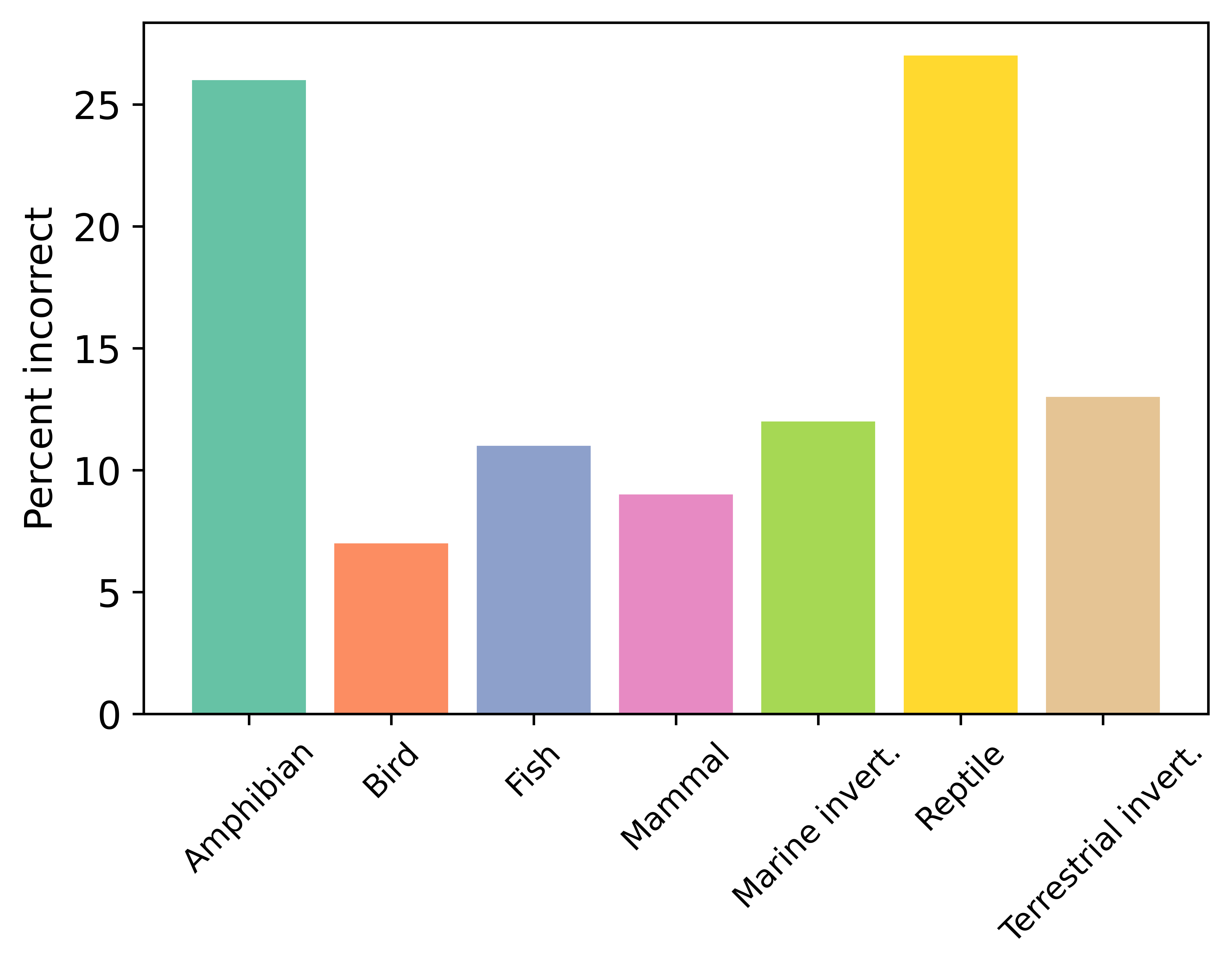}\qquad
\includegraphics[width=0.5\linewidth]{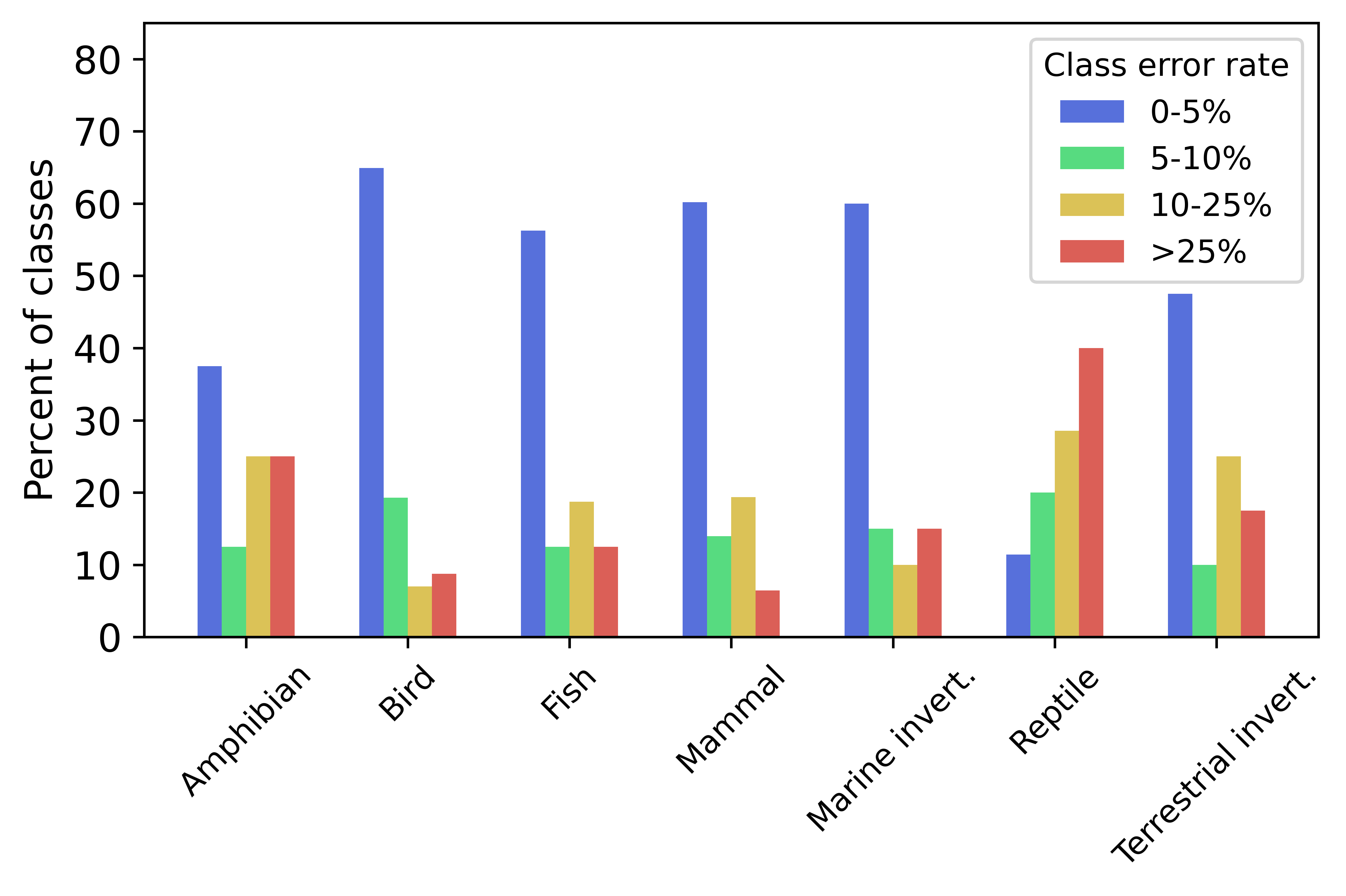}
\caption{Overall, we find that 12.3\% of wildlife images in ImageNet-1k are incorrect. Left: the percentages of incorrectly labeled images across different taxonomic groups. Right: for each taxonomic group, what percentage of the classes within that group have error rates 0-5\%, 5-10\%, 10-25\%, and $>$25\%, respectively (for example, 65\% of ImageNet-1k classes referring to birds have $<$5\% error rates, while 40\% of classes for reptiles have $>$25\% error rates).\vspace{-0.1in}}
\label{fig:incorrectID}
\end{center}
\end{figure*}

\subsection{Inaccurate Examples}

We find that 12.3 percent of the wildlife images in ImageNet-1k are incorrectly labeled and do not fall into the class in question. As shown in Figure \ref{fig:incorrectID}, error rates vary markedly between different taxonomic groups, with birds having the lowest error rate (7\%) -- consistent with the estimate of~\cite{van2015building} --  and reptiles the highest error rate (27\%). We further note that some classes have extremely high error rates; 16 classes have at least half of the images incorrect and 5 classes have more than 90\% of images incorrect (see Table \ref{table:incorrect-classes}).  As described below, we find that inaccurate labels reflect the inexperience of those originally annotating the dataset, as well as effects related to prevalence, geography, and nomenclature.

\paragraph{Annotator inexperience.} Likely the biggest factor in inaccuracies is annotator inexperience. Identification of animals is a task that in most cases requires considerable experience and expert knowledge, while the original ImageNet annotators were members of the general public without specialized expertise. Moreover, the images given to the annotators were selected through Google Image Search, which relies on textual descriptions given by the original photographers -- themselves subject to inaccuracies. There is thus the potential for two layers of annotator inexperience: first by the photographer and then by the ImageNet labeler. For instance, within the class ``crane'' (the bird), for example, 22\% percent of the images are incorrect and instead depict herons and egrets, a similar group of birds that are often confused with cranes (distressingly, many of these incorrect images show the egret \emph{Ardea alba}, which has its own class in ImageNet-1k). There appear to be a significant number of such inaccuracies for insects and other arthropods, reflecting the fact that these animals are often less familiar to the general public. In the class ``centipede,'' for example, 60\% percent of the images are incorrect, with most of the incorrect examples being images of millipedes or furry caterpillars (for comparison, centipedes are about as closely related to millipedes and caterpillars as humans are to fish).

\begin{table}[ht]
\begin{center}
\begin{tabular}{l|c}
\textbf{Class Label}                         & \textbf{Incorrect} \\ \hline
black-footed ferret, \textit{Mustela nigripes}   & 98\%    \\
rock crab, \textit{Cancer irroratus}  & 96\% \\
tailed frog, bell toad, \textit{Ascaphus trui}  & 96\%  \\
kit fox, \textit{Vulpes macrotis}  & 92\%   \\
goldfinch, \textit{Carduelis carduelis}  & 90\%   \\
green lizard, \textit{Lacerta viridis}   & 86\%  \\
night snake, \textit{Hypsiglena torquata}  & 82\%  \\
green snake, grass snake   & 70\% \\
mud turtle  & 62\% \\
horned viper, cerastes,  \textit{Cerastes cornutus} & 60\% \\
\end{tabular}\vspace{0.2in}
\caption{The top 10 classes with the highest error rates, and the percentages of incorrect images for each.\vspace{-0.1in}}
\label{table:incorrect-classes}
\end{center}
\end{table}

\paragraph{Prevalence effects.} Certain inaccuracies appear to be caused by the greater prevalence of a similar animal as compared to the class in question. For example, the ImageNet-1k class ``black-footed ferret, \textit{Mustela nigripes}'' refers to an endangered species that is uncommonly seen due to its rarity and nocturnal habits. Out of the 50 images supposedly in this class, only 2 images are correct. Almost all of the remaining 48 are photographs taken inside houses of domestic ferrets, a different species that is, not surprisingly, much more commonly seen. In the Google Image Search used to create ImageNet, it appears likely that the top search results were images of domestic ferrets that were then endorsed by the inexperienced annotators.

\paragraph{Geographic effects.} Some of the inaccuracies observed reflect geographic biases, with a species in one location having a similar name to that in another location. For example, one class refers to the goldfinch \textit{Carduelis carduelis} (a species found in Europe, Africa, and Asia) and 90\% of the images are incorrect, with most of the incorrect examples showing the American goldfinch (\textit{Spinus tristis}). It is worth noting that out of the incorrect images of birds within the dataset, our experts found that 71\% depicted birds found within the United States (see Figure \ref{fig:birdspecies}).

\paragraph{Nomenclature effects.} Other inaccuracies reflect the ways in which a technical name may be misinterpreted. The class name ``leaf beetle, chrysomelid'' for example, refers to beetles in the family Chrysomelidae. The images given for this class in ImageNet are 52\% percent incorrect, with most of the inaccurate images being other kinds of beetles sitting on leaves. In this case, many of the inaccurate images are of ladybugs, which is especially problematic since ``ladybug'' is itself a class in ImageNet.  A similar situation occurs for the class ``kit fox, \textit{Vulpes macrotis},`` which refers to a specific species of fox, and yet 92\% of the images from this class are, in fact, of young foxes from other species of fox, since ``kit'' is also the term for a young fox of any species.

\paragraph{Note on taxonomic specificity.} Out of the 269 classes referring to wildlife, our annotators determined that 130 refer to individual species (e.g.~``coho salmon''), while 139 refer to multiple species (e.g.~``scorpion''). It may be wondered whether errors are more likely to occur for the species-specific classes, as such categories are by some measure narrower. We find that indeed this is the case, but that a considerable number of errors occur in both species-specific and non-species-specific classes -- with these classes having overall error rates of 15.5\% and 9.2\%, respectively.

\subsection{Biases in Examples}

Even in cases where images are correctly classified, we nonetheless observe problematic trends in the choice of images, reflecting geographic and contextual biases. We also note numerous instances of ambiguous images.

\paragraph{Geographic bias.} According to our annotators, 51.7\% percent of the wildlife classes in ImageNet encompass more than one species. ``Jay'' for example can refer to any of a number of birds in the family Corvidae, such as the green jay (\textit{Cyanocorax luxuosus}), Eurasian jay (\textit{Garrulus glandarius}), and Sichuan jay (\textit{Perisoreus internigrans}), while ``crane'' refers to any bird in the family Gruidae. In such cases, we observe significant biases in the exact identity of the species chosen within each class. For example, all of the examples in ImageNet-1k for the class ``jay'' are correct in that they depict jays. However, 62\% of these images show the blue jay (\textit{Cyanocitta cristata}), which is common in the U.S.~but is only one of 49 species of jay worldwide. Once again, such biases likely result from the Google Image Search methodology originally used to create ImageNet; it is well-known that certain geographies are better represented in web search results, especially for English-language searches.

Overall, our annotators analyzed all 32 non-specific classes of birds in the ImageNet dataset and found an average of 52\% of the images were of species present in the United States (see Figure \ref{fig:birdspecies}). In some cases where large numbers of species outside the U.S.~were included (for example in the ``crane'' and ``macaw'' classes), many images appear to be of birds in zoos or as pets, and therefore could still represent photos taken in the U.S.

\paragraph{Contextual bias.} The context in which species are depicted within the dataset is also subject to significant bias, e.g.~many of the photographs of jellyfish seemingly being taken inside aquariums. Contextual bias is particularly blatant for certain classes referring to fish, in which fishing scenarios (e.g.~the fish being held out of the water by a human) vastly outweigh images of fish in their natural environment. Across the classes for tench, barracouta, coho salmon, sturgeon, and gar, 61.6\% percent of the images include humans.  This bias is also present in mammals such as the hartebeest that are often hunted for game, resulting in images of dead animals accompanied by the humans who have shot them. Some contextual bias is of course inevitable in image-taking (e.g., butterflies may be more likely to be photographed while near the ground, rather than at the top of a tree), but such a significant and avoidable bias is noteworthy.

\begin{table}[t]
\begin{center}
{\small
\begin{tabular}{l|ccccc}
\textbf{Group}                 & \textbf{Blur} & \multicolumn{1}{l}{\textbf{Collage}} & \textbf{Fake} & \textbf{Human} & \textbf{Multi} \\ \hline
Amphibian& 0.3\% & 0.3\%    & 0.8\%    & 6.0\%  & 0.8\%           \\
Bird& 0.4 \% & 0.4\%      & 0.4 \%     & 1.0\%   & 3.2\%           \\
Fish& 10.8\% & 0.1\%      & 1.3\%    & 24.6\%    & 13.3\%      \\
Mammal& 2.7\%  & 1.3\%       & 2.0\%     & 6.3\%    & 2.4\%          \\
Marine Invert.& 4.4\%  & 0.2\%      & 3.7\%      & 14.3\%  & 22.6\%           \\
Reptile& 0.9\% & 0.2\%      & 0.9\%     & 9.5\%    & 2.1\%  \\ 
Terr.~Invert.& 	0.5\%  & 0.3\%  & 1.1\%      & 2.4\%    & 1.4 \%          \\\hline   Overall & 2.2\% & 0.6\% & 1.4\% & 6.7\% & 4.5 \%
\end{tabular}
}
\end{center}
\caption{Percentages of images with various confusing attributes: respectively, severe blur, a collage of multiple images, depiction of a fake animal such as a painting or stuffed toy, the presence of humans, and the presence of multiple species of animals in the same image.\vspace{-0.1in}}
\label{table:image-annotations}
\end{table}

\paragraph{Bad or ambiguous images.} In addition to the biases noted above, many of the images were confusing for other reasons. As described in Table \ref{table:image-annotations}, our annotators observed 6.7\% of images had humans present (including but not limited to the fishing images described above), 4.5\% of images had multiple (non-human) species present in the image, 1.4\% of images showed an artificial representation of the animal in question (such as an illustration, toy or sculpture), and 0.6\% of images were actually multiple images merged into a collage. The annotators also noted that 2.2\% of images were too blurry or low quality to allow for reliable identification. See Figure~\ref{fig:examples} for examples of these failure modes.

\section{Discussion} \label{sec:discussion}

\begin{figure}[ht]
    \centering
    \includegraphics[width=0.24\linewidth]{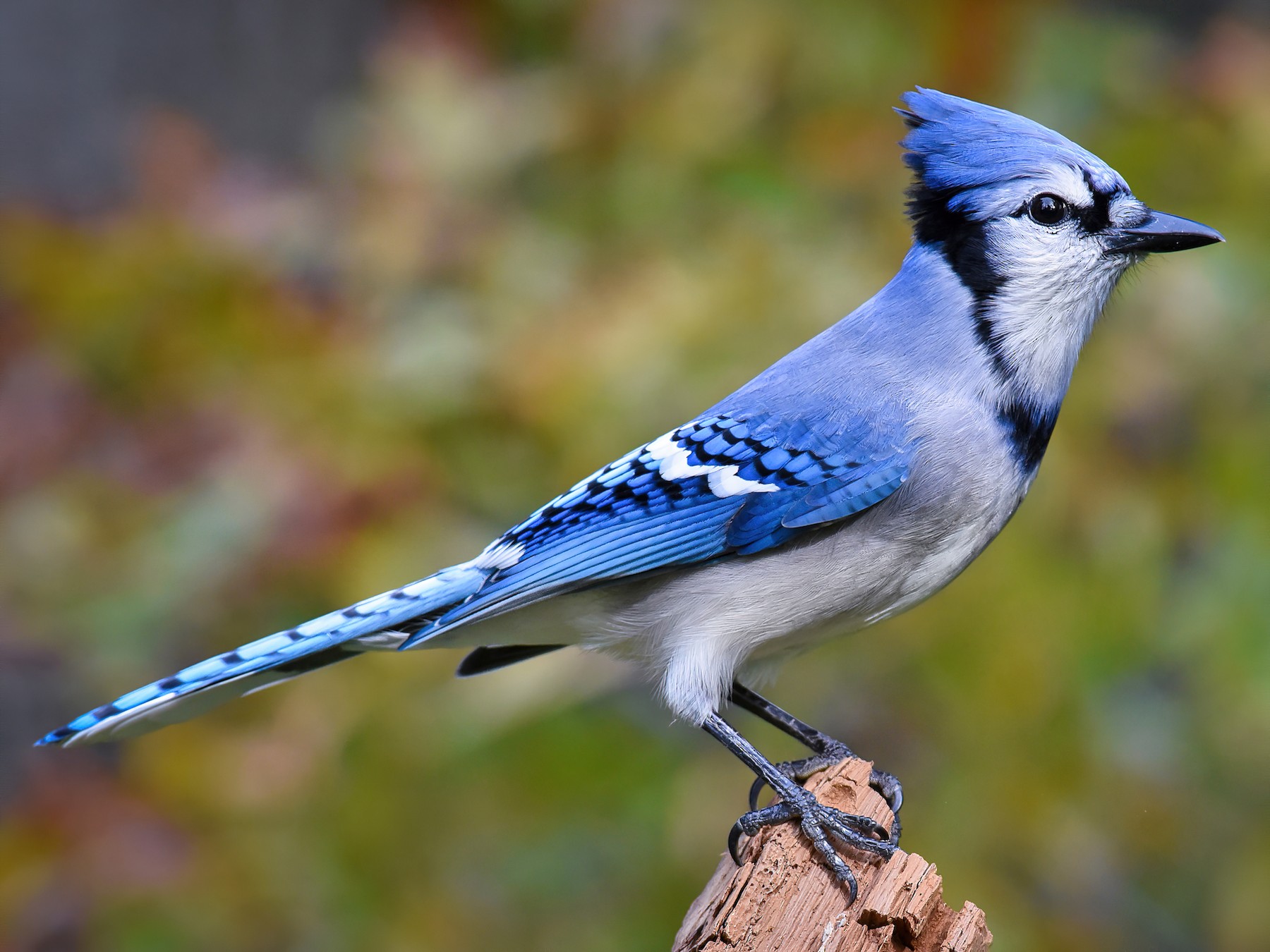}
    \includegraphics[width=0.24\linewidth]{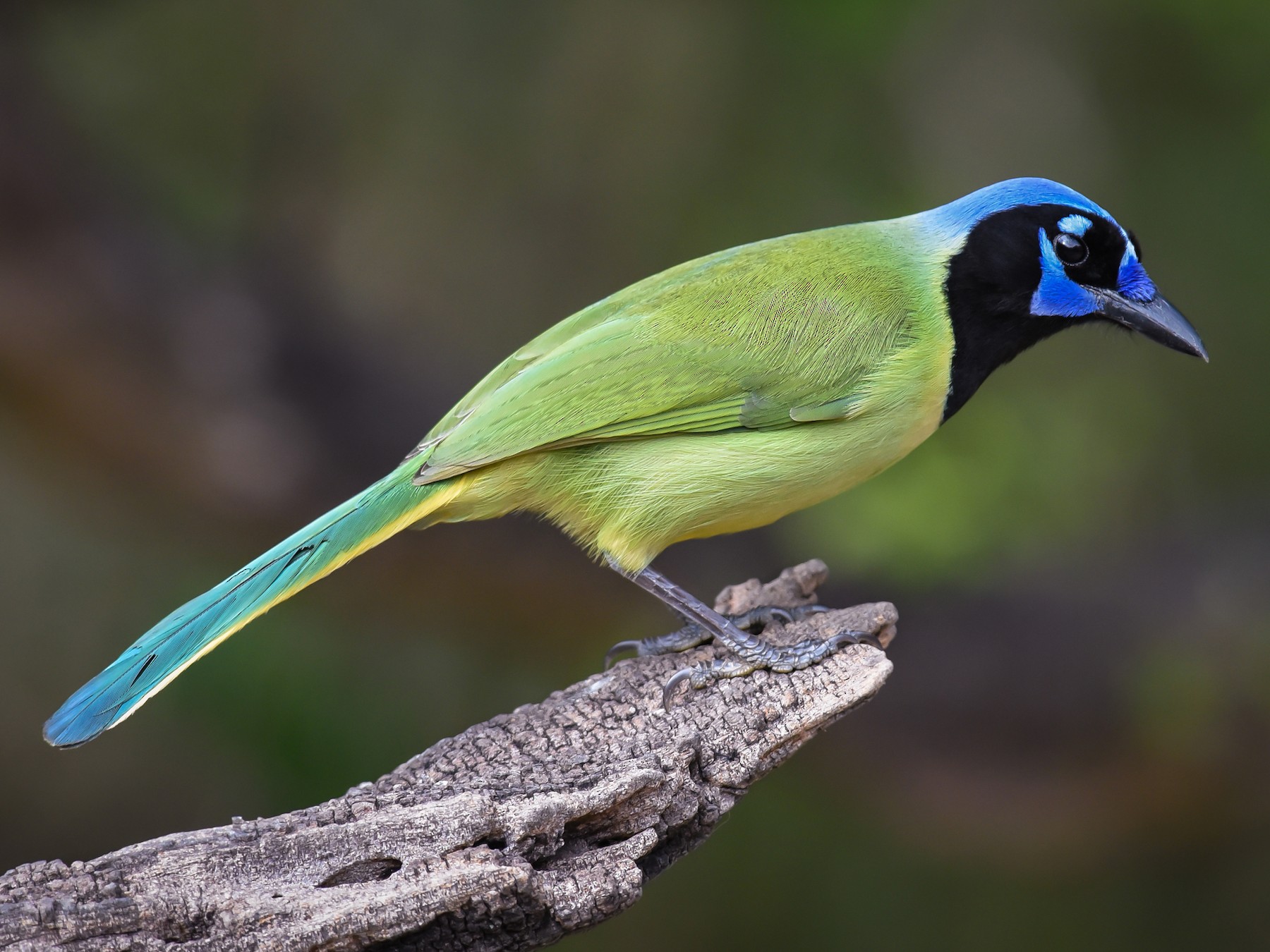}
    \includegraphics[width=0.24\linewidth]{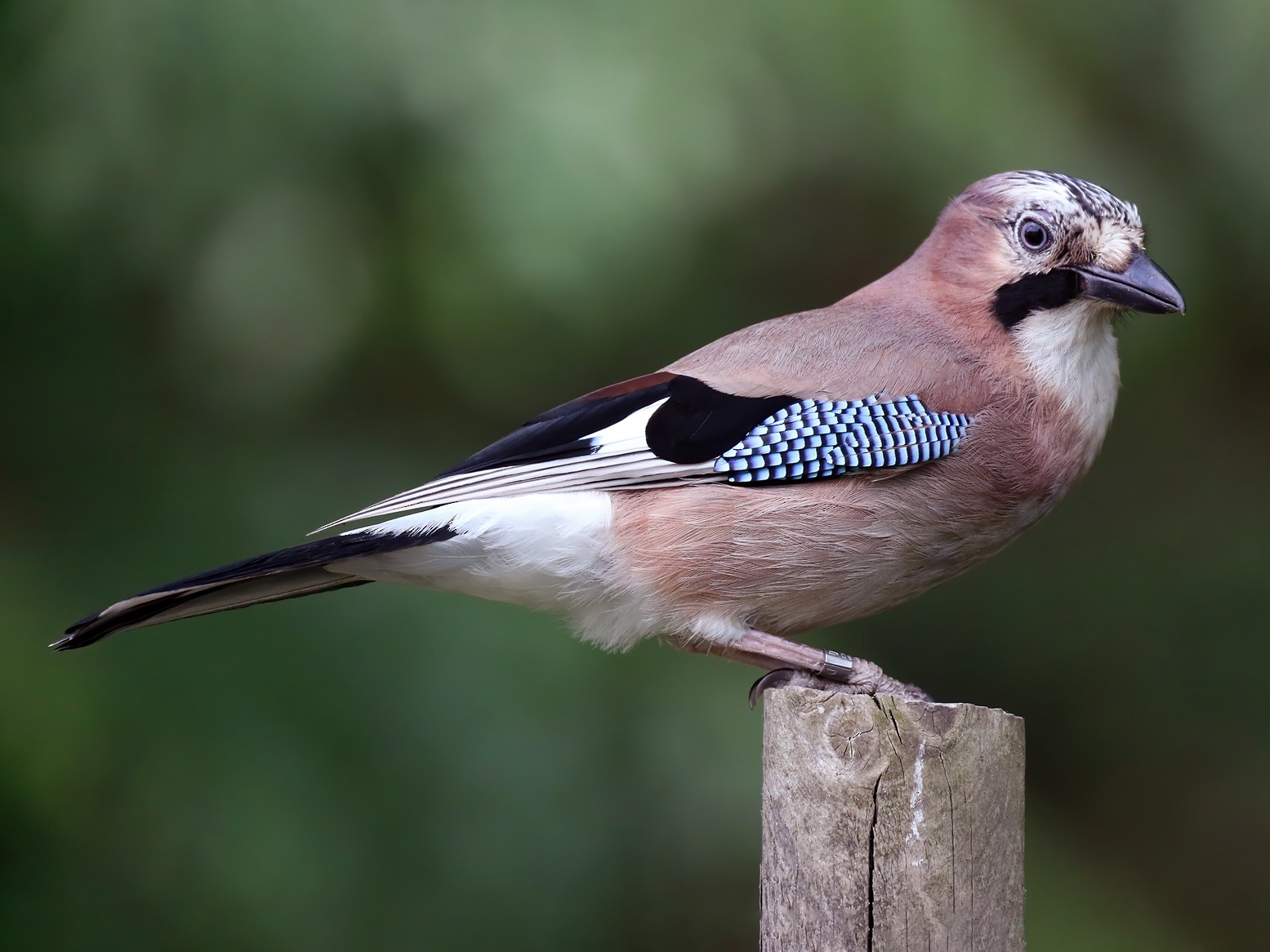}
    \includegraphics[width=0.24\linewidth]{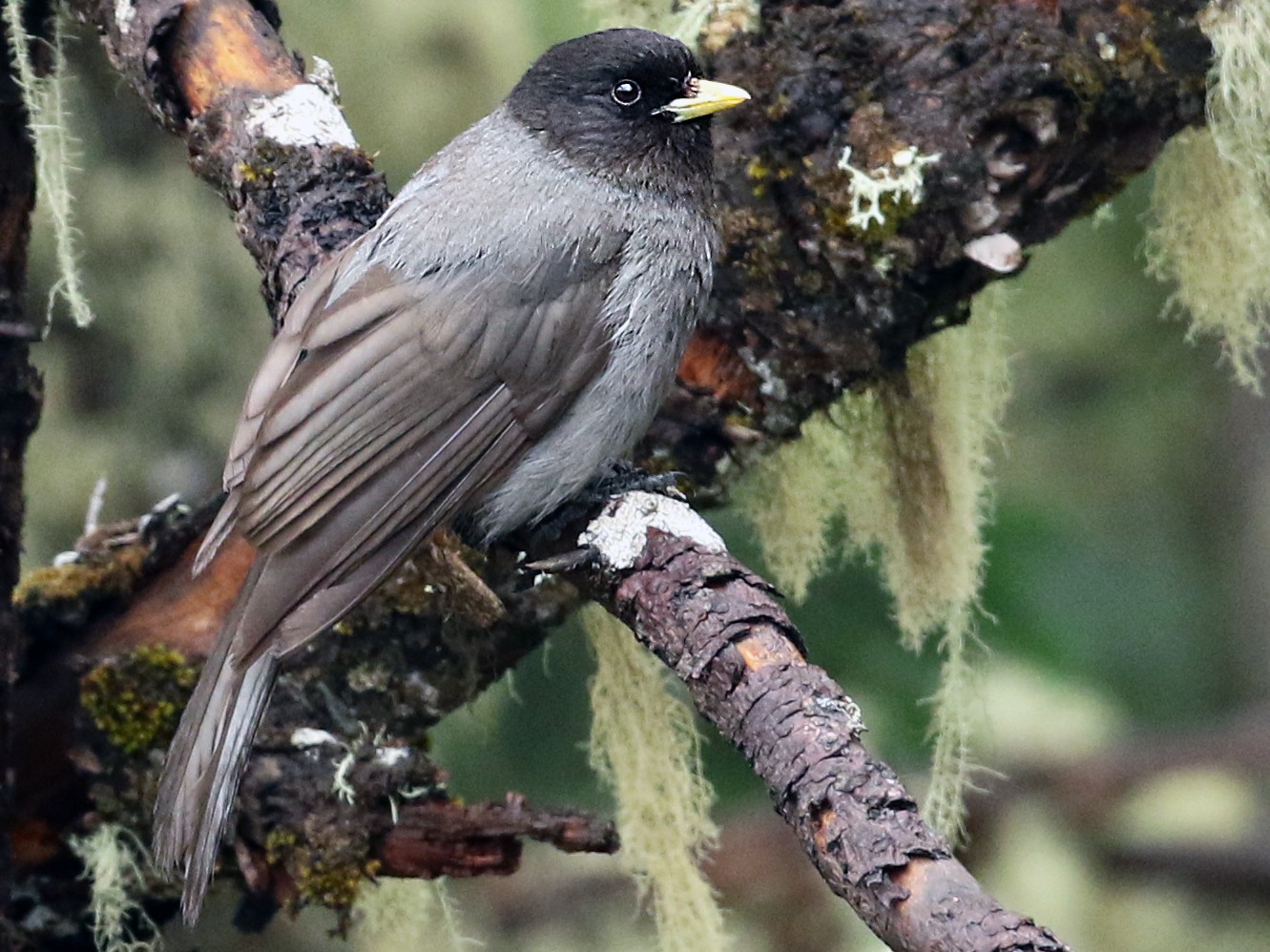}
    \caption{Four of the 49 species of jays found around the world: the blue jay, green jay, Eurasian jay, and Sichuan jay. Despite the visual and geographic diversity of jays, 62\% of the ``jay'' images in ImageNet are of the blue jay (a species found only in the U.S.~and Canada). This example illustrates geographic inequities as well as the fact that ImageNet classes are implicitly made up of imbalanced subclasses requiring few-shot (or zero-shot) generalization. Image credit: see Macaulay Library 84878861, 79048551, 241250451, 41284331.\vspace{-0.1in}}
    \label{fig:jays}
\end{figure}

Our findings have many implications for the machine learning community. First, they call into question the ways in which performance on ImageNet-1k is used as a measure of success. It continues to be common practice for ML models to be benchmarked on ImageNet-1k, and reviewers frequently request this. Implicit in such tests are two assumptions that we show to be false:
\begin{itemize}
    \item Validation accuracy is supposed to reflect how well the model is learning. In fact, we find that, within wildlife images, the validation set is 12.3\% percent incorrectly labeled. Many of these examples should be correctly labeled as another class in ImageNet (e.g.~images labeled as ``crane'' that are actually ``egret''), while others are ambiguous (is it an ``African elephant'' or a ``tusker''?), and still others are out-of-distribution examples without any correct ImageNet label. Notably, none of these inaccuracies can be dismissed as simply random noise -- instead, they admit complex correlations across classes, yielding hard-to-predict effects on overall accuracy.
    \item ImageNet is supposed to have balanced classes. In fact, we find that many classes are implicitly made up of subclasses that are unevenly represented. For example, a model trained on the ``jay'' class, where 62\% of the images show a single species of jay, may generalize poorly to the other 48 species of jays in the world, many of which look radically different (see Figure \ref{fig:jays}).  It is likely that in many cases, such imbalances result in validation and test examples that are highly dissimilar from all or most examples in the training set, essentially requiring few- or zero-shot generalization.
\end{itemize}
Failure of these basic assumptions further calls into question the utility of ImageNet as a benchmark and the validity of state-of-the-art claims, especially given the minute margins by which state-of-the-art is established on ImageNet.

Moreover, while there are many wildlife-focused datasets such as iNaturalist, iWildCam, and NABirds, the ImageNet dataset is still used in various contexts where its performance on animal identification matters. It is often used to pretrain networks for fine-tuning on wildlife recognition datasets, especially by researchers who do not have access to the computational resources required to train models from scratch. Inaccuracies and biases in the original ImageNet dataset can therefore directly affect the performance and biases of downstream models. Furthermore, ImageNet is also used in generic contexts such as classifying photos and image search. This can lead to the propagation and magnification of biases if classification algorithms influence the animals that people learn or the results that appear in search engines. For example, if most of the ImageNet images of ``jay'' are dominated by a small subset of the possible species, then search results may miss relevant information or be implicitly biased towards content from certain geographies.

Finally, our work calls attention to broader issues in the machine learning ecosystem. The biases and inaccuracies we identify within wildlife imagery are, of course, not unique to wildlife data. Rather, the objective nature of taxonomic classification can make it easier to observe and quantify issues that likely arise extensively within other types of data (see e.g.~\cite{crawford2019excavating,prabhu2020large}). Nor have ImageNet's failures prevented similar weaknesses in more recently developed datasets. Machine learning continues to draw from large Internet corpora with significant biases, using annotators who often lack relevant domain expertise. Tasks and classes within a dataset may likewise be chosen by individuals without relevant expertise, and who may be biased in their geographies or lived experiences.

\section{Conclusion} \label{sec:conclusion}

In this work, we analyze how biodiversity is (mis)represented in the ImageNet-1k dataset, working with domain experts in order to evaluate both the soundness of class definitions and the quality of the images that are included in each class. Our results show extensive biases, both geographic and cultural, in the choice of wildlife represented in the dataset, as well as high incidence of unclear class definitions and overlap between classes. We also find a high percentage of incorrect labels, with some classes almost completely incorrect. Even correctly classified images are frequently confusing -- including the presence of artificial animals, humans, or other animal species -- and the images chosen to represent classes are strongly biased towards the United States and Europe.

We wish that our work offered a quick fix for the many failings we have identified in ImageNet-1k, but unfortunately solutions ultimately require larger-scale, indeed systemic, change. It is not sufficient simply to relabel all inaccurate images we have found -- for many of them, the true species falls outside any ImageNet class, and such examples would need to be replaced by new images. In some cases, as we have seen, the majority of the class would need to be replaced. More insidious still are the biases present in the images and the biases and inconsistencies in the class definitions, which would essentially require replacing the structure of the whole dataset. We hope to see future work quantifying how the perceived efficacy of different ML models has been skewed by errors and biases in ImageNet-1k. We also look forward to improved standards in ML for the participation of diverse domain experts in the creation and annotation of datasets, both with relevance to biodiversity and more broadly.

\section*{Ethical statement}
The goal of our work is to quantify the extent to which the natural world, namely wildlife, is under- and mis-represented in ImageNet-1k. We consider the ethical and practical implications that these biases and inaccuracies pose for the development and use of machine learning algorithms. With respect to the methods used in the study, we endeavored to follow best practices in responsible annotation of classes and images. While the initial ImageNet dataset was crowdsourced using the Amazon Mechanical Turk platform, there has since been extensive work that has questioned the ethics of anonymous crowdsourcing platforms~\cite{irani2016stories,shmueli2021beyond}. Given these reasons, as well as the importance of domain expertise for our analysis, we recruited annotators ourselves, taking into account characteristics such as the location, seniority and domain of expertise of each annotators. Furthermore, all of our annotators were remunerated at a fixed rate per image annotated, intended to approximate an hourly rate of \$50 per hour, with the final total annotation cost approaching \$10,000.

\section*{Acknowledgments}
We would like to thank all of our expert annotators for their contributions: Kerri Lynn Ackerly, Ashton Asbury, Hazel Byrne, Michele Chiacchio, Eamon C.~Corbett, Terrence Demos, Jessica Eggers, Miquéias Ferrão, Jeff Goddard,  Alex Harman, Martin Hauser,  Bob Jacobs, Daniel Kluza, Ashwini V.~Mohan, Claude Nozères, Kris Sabbi, Subir B.~Shakya, Jacob Socolar, David Vander Pluym, and Tyus D.~Williams. This work was supported in part by the Canada CIFAR AI Chairs program. 

\bibliographystyle{plainnat}

\begin{thebibliography}{10}

\bibitem{barbu2019objectnet}
Andrei Barbu, David Mayo, Julian Alverio, William Luo, Christopher Wang, Dan
  Gutfreund, Josh Tenenbaum, and Boris Katz.
\newblock Objectnet: A large-scale bias-controlled dataset for pushing the
  limits of object recognition models.
\newblock {\em Advances in Neural Information Processing Systems (NeurIPS)},
  32, 2019.

\bibitem{barz2020we}
Bj{\"o}rn Barz and Joachim Denzler.
\newblock Do we train on test data? purging {CIFAR} of near-duplicates.
\newblock {\em Journal of Imaging}, 6(6):41, 2020.

\bibitem{beery2021iwildcam}
Sara Beery, Arushi Agarwal, Elijah Cole, and Vighnesh Birodkar.
\newblock The {iWildCam} 2021 competition dataset.
\newblock {\em Preprint arXiv:2105.03494}, 2021.

\bibitem{beery2019iwildcam}
Sara Beery, Grant Van~Horn, Oisin Mac~Aodha, and Pietro Perona.
\newblock The {iWildcam} 2018 challenge dataset.
\newblock {\em Preprint arXiv:1904.05986}, 2019.

\bibitem{beery2018recognition}
Sara Beery, Grant Van~Horn, and Pietro Perona.
\newblock Recognition in terra incognita.
\newblock In {\em Proceedings of the European conference on computer vision
  (ECCV)}, pages 456--473, 2018.

\bibitem{beyer2020we}
Lucas Beyer, Olivier~J H{\'e}naff, Alexander Kolesnikov, Xiaohua Zhai, and
  A{\"a}ron van~den Oord.
\newblock Are we done with {ImageNet}?
\newblock {\em Preprint arXiv:2006.07159}, 2020.

\bibitem{crawford2019excavating}
Kate Crawford and Trevor Paglen.
\newblock Excavating {AI}: The politics of images in machine learning training
  sets.
\newblock {\em AI and Society}, 2019.

\bibitem{deng2009imagenet}
Jia Deng, Wei Dong, Richard Socher, Li-Jia Li, Kai Li, and Li~Fei-Fei.
\newblock {ImageNet}: A large-scale hierarchical image database.
\newblock In {\em IEEE Conference on Computer Vision and Pattern Recognition
  (CVPR)}, pages 248--255, 2009.

\bibitem{denton2020bringing}
Emily Denton, Alex Hanna, Razvan Amironesei, Andrew Smart, Hilary Nicole, and
  Morgan~Klaus Scheuerman.
\newblock Bringing the people back in: Contesting benchmark machine learning
  datasets.
\newblock {\em Preprint arXiv:2007.07399}, 2020.

\bibitem{seek}
iNaturalist.
\newblock Seek.
\newblock \url{https://www.inaturalist.org/pages/seek_app}, 2022.

\bibitem{irani2016stories}
Lilly~C Irani and M~Six Silberman.
\newblock Stories we tell about labor: Turkopticon and the trouble with"
  design".
\newblock In {\em Proceedings of the 2016 CHI conference on human factors in
  computing systems}, pages 4573--4586, 2016.

\bibitem{koch2021reduced}
Bernard Koch, Emily Denton, Alex Hanna, and Jacob~G Foster.
\newblock Reduced, reused and recycled: The life of a dataset in machine
  learning research.
\newblock In {\em Conference on Neural Information Processing Systems
  (NeurIPS)}, 2021.

\bibitem{avibase}
Denis Lepage.
\newblock Avibase.
\newblock \url{https://avibase.bsc-eoc.org/}, 2022.

\bibitem{lin2014microsoft}
Tsung-Yi Lin, Michael Maire, Serge Belongie, James Hays, Pietro Perona, Deva
  Ramanan, Piotr Doll{\'a}r, and C~Lawrence Zitnick.
\newblock Microsoft {COCO}: Common objects in context.
\newblock In {\em European conference on computer vision}, pages 740--755.
  Springer, 2014.

\bibitem{miller1995wordnet}
George~A Miller.
\newblock {WordNet}: a lexical database for {E}nglish.
\newblock {\em Communications of the ACM}, 38(11):39--41, 1995.

\bibitem{mora2011many}
Camilo Mora, Derek~P Tittensor, Sina Adl, Alastair~GB Simpson, and Boris Worm.
\newblock How many species are there on {Earth} and in the ocean?
\newblock {\em PLoS biology}, 9(8):e1001127, 2011.

\bibitem{northcutt2021pervasive}
Curtis~G Northcutt, Anish Athalye, and Jonas Mueller.
\newblock Pervasive label errors in test sets destabilize machine learning
  benchmarks.
\newblock In {\em Conference on Neural Information Processing Systems
  (NeurIPS)}, 2021.

\bibitem{prabhu2020large}
Vinay~Uday Prabhu and Abeba Birhane.
\newblock Large image datasets: A pyrrhic win for computer vision?
\newblock In {\em IEEE Winter Conference on Applications of Computer Vision
  (WACV)}, pages 1536--1546, 2021.

\bibitem{recht2019imagenet}
Benjamin Recht, Rebecca Roelofs, Ludwig Schmidt, and Vaishaal Shankar.
\newblock Do {ImageNet} classifiers generalize to {ImageNet}?
\newblock In {\em International Conference on Machine Learning (ICML)}, pages
  5389--5400, 2019.

\bibitem{russakovsky2015imagenet}
Olga Russakovsky, Jia Deng, Hao Su, Jonathan Krause, Sanjeev Satheesh, Sean Ma,
  Zhiheng Huang, Andrej Karpathy, Aditya Khosla, Michael Bernstein, et~al.
\newblock {ImageNet} large scale visual recognition challenge.
\newblock {\em International Journal of Computer Vision}, 115(3):211--252,
  2015.

\bibitem{scheuerman2021datasets}
Morgan~Klaus Scheuerman, Alex Hanna, and Emily Denton.
\newblock Do datasets have politics? disciplinary values in computer vision
  dataset development.
\newblock {\em Proceedings of the ACM on Human-Computer Interaction},
  5(CSCW2):1--37, 2021.

\bibitem{shmueli2021beyond}
Boaz Shmueli, Jan Fell, Soumya Ray, and Lun-Wei Ku.
\newblock Beyond fair pay: Ethical implications of nlp crowdsourcing.
\newblock {\em arXiv preprint arXiv:2104.10097}, 2021.

\bibitem{sullivan2009ebird}
Brian~L Sullivan, Christopher~L Wood, Marshall~J Iliff, Rick~E Bonney, Daniel
  Fink, and Steve Kelling.
\newblock {eBird}: A citizen-based bird observation network in the biological
  sciences.
\newblock {\em Biological conservation}, 142(10):2282--2292, 2009.

\bibitem{tsipras2020imagenet}
Dimitris Tsipras, Shibani Santurkar, Logan Engstrom, Andrew Ilyas, and
  Aleksander Madry.
\newblock From {ImageNet} to image classification: Contextualizing progress on
  benchmarks.
\newblock In {\em International Conference on Machine Learning (ICML)}, pages
  9625--9635, 2020.

\bibitem{van2015building}
Grant Van~Horn, Steve Branson, Ryan Farrell, Scott Haber, Jessie Barry, Panos
  Ipeirotis, Pietro Perona, and Serge Belongie.
\newblock Building a bird recognition app and large scale dataset with citizen
  scientists: The fine print in fine-grained dataset collection.
\newblock In {\em Proceedings of the IEEE Conference on Computer Vision and
  Pattern Recognition (CVPR)}, pages 595--604, 2015.

\bibitem{van2018inaturalist}
Grant Van~Horn, Oisin Mac~Aodha, Yang Song, Yin Cui, Chen Sun, Alex Shepard,
  Hartwig Adam, Pietro Perona, and Serge Belongie.
\newblock The {iNaturalist} species classification and detection dataset.
\newblock In {\em Proceedings of the IEEE Conference on Computer Vision and
  Pattern Recognition (CVPR)}, pages 8769--8778, 2018.

\bibitem{van2017devil}
Grant Van~Horn and Pietro Perona.
\newblock The devil is in the tails: Fine-grained classification in the wild.
\newblock {\em Preprint arXiv:1709.01450}, 2017.

\bibitem{yang2020towards}
Kaiyu Yang, Klint Qinami, Li~Fei-Fei, Jia Deng, and Olga Russakovsky.
\newblock Towards fairer datasets: Filtering and balancing the distribution of
  the people subtree in the {ImageNet} hierarchy.
\newblock In {\em Proceedings of the 2020 Conference on Fairness,
  Accountability, and Transparency (FAccT)}, pages 547--558, 2020.

\bibitem{yang2021study}
Kaiyu Yang, Jacqueline Yau, Li~Fei-Fei, Jia Deng, and Olga Russakovsky.
\newblock A study of face obfuscation in {ImageNet}.
\newblock In {\em International Conference on Machine Learning (ICML)}, 2022.

\end{thebibliography}

\appendix

\clearpage

\section{Annotation Details} 
\label{app:annotation}
We divided ImageNet-1k wildlife classes into sets according to standard domains of expertise for ecologists and taxonomists: birds, fish, herps (including amphibians and reptiles), marine invertebrates, primates, other mammals, terrestrial invertebrates. We further subdivided birds into sets A and B since there were a large number of bird classes. For each set of classes, we recruited expert annotators with at least graduate-level experience in the relevant taxonomic group. Labels were completed according by the following numbers of annotators:
\begin{table}[ht]
    \centering
    \begin{tabular}{c|c}
         ImageNet-1k Classes & \# Annotators \\\hline
         Birds A & 3\\
         Birds B & 1 \\
         Fish & 2 \\
         Herps & 3 \\
         Marine Inverts.&3\\
         Primates & 3\\
         Other Mammals & 2\\
         Terrestrial Inverts.& 3
    \end{tabular}
\end{table}

Each annotator answered questions on all classes within their set and all images belonging to those classes, with the various annotators completing their work independently without access to each others' answers. To obtain final results, we obtained a consensus for each question, based on whether a majority of annotators answered Yes or No (ties were broken in favor of Yes answers, e.g.~if there was doubt as to whether an image was correctly identified or not, the tie was broken in favor of the image being correct). Consensus between annotators was high, with a 0.72 average Krippendorff alpha value.

It is worth noting that annotators, as experts in wildlife identification, were in general highly invested in the success of the project and extremely thorough in their work. Many wrote lengthy messages to us explaining their logic -- for example, the following message about the ``cricket'' class:

\begin{quote}
``Crickets'' \emph{sensu lato} encompasses suborder Ensifera, which is essentially any orthopteran that isn't a grasshopper.

Ensifera is comprised of two infraorders: Gryllidea (which includes the so-called ``true crickets'' (family Gryllidae)) and Tettigoniidea.

The vast majority of families in Tettigoniidea are referred to as crickets, however one exception is family Tettigoniidae -- katydids. That said, Tettigoniids can be referred to as ``bush-cricket'' (e.g., in the UK) rather than ``katydids''.

In my own nomenclatorial/classification world view, I don't regard katydids as crickets \emph{per se}. When I think ``cricket'' I think family Gryllidae, and this was my initial approach to the cricket images.

When it comes to AI image classification, however, calling a katydid a cricket isn't necessarily incorrect. It's going to depend upon where the classification boundary is set (e.g.,~suborder Ensifera (non-grasshopper orthopterans) vs. family Gryllidae (``true crickets'')) or how the classification is defined (e.g., non-grasshopper orthopterans referred to as crickets within a vernacular).
\end{quote}

In addition to the class-specific and image-specific questions, we obtained species-level identifications of all images in the bird classes from one annotator in each of Birds A and Birds B. These annotators were able to provide species-level identifications for 86.8\% of the images (the other images either did not show relevant diagnostic features or the birds in question were not identifiable to species level from images alone). The bird annotators were extremely diligent in this work, even providing species-level identifications for a ground squirrel and hawkmoth that were erroneously classified as birds by ImageNet-1k.
\end{document}